\documentclass[10pt,twocolumn,letterpaper]{article}

\usepackage{iccv}
\usepackage{times}
\usepackage{epsfig}
\usepackage{graphicx}
\usepackage{amsmath}
\usepackage{amssymb}

\usepackage{subcaption}
\usepackage{multirow}
\usepackage[hyphens]{url}
\usepackage{caption}
\usepackage[breaklinks=true,colorlinks,bookmarks=false]{hyperref}

\iccvfinalcopy 

\def\iccvPaperID{****} 

\ificcvfinal\fi

\begin{document}
	
	\title{Gaussian YOLOv3: An Accurate and Fast Object Detector Using Localization Uncertainty for Autonomous Driving}
	
	\author{Jiwoong Choi\textsuperscript{1},~Dayoung Chun\textsuperscript{1},~Hyun Kim\textsuperscript{2},~and~Hyuk-Jae Lee\textsuperscript{1}\\
		\textsuperscript{1}Seoul National University,~\textsuperscript{2}Seoul National University of Science and Technology\\
		{\tt\small {\{jwchoi,~jjeonda\}@capp.snu.ac.kr}}, {\tt\small hyunkim@seoultech.ac.kr}, {\tt\small hjlee@capp.snu.ac.kr}
	}
	
	\maketitle
	\ificcvfinal\thispagestyle{empty}\fi
	
	\begin{abstract}
		The use of object detection algorithms is becoming increasingly important in autonomous vehicles, and object detection at high accuracy and a fast inference speed is essential for safe autonomous driving. A false positive (FP) from a false localization during autonomous driving can lead to fatal accidents and hinder safe and efficient driving. Therefore, a detection algorithm that can cope with mislocalizations is required in autonomous driving applications. This paper proposes a method for improving the detection accuracy while supporting a real-time operation by modeling the bounding box (bbox) of YOLOv3, which is the most representative of one-stage detectors, with a Gaussian parameter and redesigning the loss function. In addition, this paper proposes a method for predicting the localization uncertainty that indicates the reliability of bbox. By using the predicted localization uncertainty during the detection process, the proposed schemes can significantly reduce the FP and increase the true positive (TP), thereby improving the accuracy. Compared to a conventional YOLOv3, the proposed algorithm, Gaussian YOLOv3, improves the mean average precision (mAP) by 3.09 and 3.5 on the KITTI and Berkeley deep drive (BDD) datasets, respectively. Nevertheless, the proposed algorithm is capable of real-time detection at faster than 42 frames per second (fps) and shows a higher accuracy than previous approaches with a similar fps. Therefore, the proposed algorithm is the most suitable for autonomous driving applications.
	\end{abstract}
	
	\section{Introduction}
	
	In recent years, deep learning has been actively applied in various fields including computer vision~\cite{he2016deep}, autonomous driving~\cite{dai2019hybridnet}, and social network services~\cite{liu2013deep}. The development of sensors and GPU along with deep learning algorithms has accelerated research into autonomous vehicles based on artificial intelligence. An autonomous vehicle with self-driving capability without a driver intervention must accurately detect cars, pedestrians, traffic signs, traffic lights, etc. in real time to ensure safe and correct control decisions~\cite{wu2017squeezedet}. To detect such objects, various sensors such as cameras, light detection and ranging (Lidar), and radio detection and ranging (Radar) are generally used in autonomous vehicles~\cite{zhang2014roadview}. Among these various types of sensors, a camera sensor can accurately identify the object type based on texture and color features and is more cost-effective~\cite{wei2013towards} than other sensors. In particular, deep-learning based object detection using camera sensors is becoming more important in autonomous vehicles because it achieves a better level of accuracy than humans in terms of object detection, and consequently it has become an essential method~\cite{hu2019sinet} in autonomous driving systems.
	
	An object detection algorithm for autonomous vehicles should satisfy the following two conditions. First, a high detection accuracy of the road objects is required. Second, a real-time detection speed is essential for a rapid response of a vehicle controller and a reduced latency. Deep-learning based object detection algorithms, which are indispensable in autonomous vehicles, can be classified into two categories: two-stage and one-stage detectors. Two-stage detectors,~\eg, Fast R-CNN~\cite{girshick2015fast}, Faster R-CNN~\cite{ren2015faster}, and R-FCN~\cite{dai2016r}, conduct a first stage of region proposal generation, followed by a second stage of object classification and bbox regression. These methods generally show a high accuracy but have a disadvantage of a slow detection speed and lower efficiency. One-stage detectors,~\eg, SSD~\cite{liu2016ssd} and YOLO~\cite{redmon2016you}, conduct object classification and bbox regression concurrently without a region proposal stage. These methods generally have a fast detection speed and high efficiency but a low accuracy. In recent years, to take advantage of both types of method and to compensate for their respective disadvantages, object detectors combining various schemes have been widely studied~\cite{cai2016unified,hu2019sinet,zhao2018cfenet,zhang2018single,liu2018receptive}. MS-CNN~\cite{cai2016unified}, a two-stage detector, improves the detection speed by conducting detection on various intermediate network layers. SINet~\cite{hu2019sinet}, also a two-stage detector, enables a fast detection using a scale-insensitive network. CFENet~\cite{zhao2018cfenet}, a one-stage detector, uses a comprehensive feature enhancement module based on SSD to improve the detection accuracy. RefineDet~\cite{zhang2018single}, also a one-stage detector, improves the detection accuracy by applying an anchor refinement module and an object detection module. Another one-stage detector, RFBNet~\cite{liu2018receptive}, applies a receptive field block to improve the accuracy. However, using an input resolution of 512 $\times$ 512 or higher, which is widely applied in object detection algorithms for achieving a high accuracy, previous studies~\cite{cai2016unified,hu2019sinet,zhao2018cfenet,zhang2018single} have been unable to meet a real-time detection speed of above 30 fps, which is a prerequisite for self-driving applications. Even if real-time detection is possible in~\cite{liu2018receptive}, it is difficult to apply to autonomous driving due to a low accuracy. This indicates that these previous schemes are incomplete in terms of a trade-off between accuracy and detection speed, and consequently, have a limitation in their application to self-driving systems.
	
	In addition, one of the most critical problems of most conventional deep-learning based object detection algorithms is that, whereas the bbox coordinates (\textit{i.e.}, localization) of the detected object are known, the uncertainty of the bbox result is not. Thus, conventional object detectors cannot prevent mislocalizations (\textit{i.e.}, FPs) because they output the deterministic results of the bbox without information regarding the uncertainty. In autonomous driving, an FP denotes an incorrect detection result of bbox on an object that is not the ground-truth (GT), or an inaccurate detection result of bbox on the GT, whereas a TP denotes an accurate detection result of bbox on the GT. An FP is extremely dangerous under autonomous driving because it causes excessive reactions such as unexpected braking, which can reduce the stability and efficiency of driving and lead to a fatal accident~\cite{fp_cite,seo2009self} as well as confusion in the determination of an accurate object detection. In other words, it is extremely important to predict the uncertainty of the detected bboxes and to consider this factor along with the objectness score and class scores for reducing the FP and preventing autonomous driving accidents. For this reason, various studies have been conducted on predicting uncertainty in deep learning. Kendall~\etal~\cite{kendall2017uncertainties} proposed a modeling method for uncertainty prediction using a Bayesian neural network in deep learning. Feng~\etal~\cite{feng2018towards} proposed a method for predicting uncertainty by applying Kendall~\etal’s scheme~\cite{kendall2017uncertainties} to 3D vehicle detection using a Lidar sensor. However, the methods proposed by Kendall~\etal~\cite{kendall2017uncertainties} and Feng~\etal~\cite{feng2018towards} only predict the level of uncertainty, and do not utilize this factor in actual applications. Choi~\etal~\cite{choi2018uncertainty} proposed a method for predicting uncertainty in real time using a Gaussian mixture model and applied the method to an autonomous driving application. However, it was applied to the steering angle, and not object detection, and a complicated distribution is therefore modeled, increasing the computational complexity. He~\etal~\cite{he2018softer} proposed an approach for predicting uncertainty and utilized it toward object detection. However, because they focused on a two-stage detector, their method cannot support a real-time operation, and remaining a bbox overlap problem, so it is unsuitable for self-driving applications.
	\begin{figure*}[t!]
		\centering
		\hspace{-1cm}
		\begin{subfigure}[t]{0.6\textwidth}
			\centering
			\includegraphics[scale=0.82]{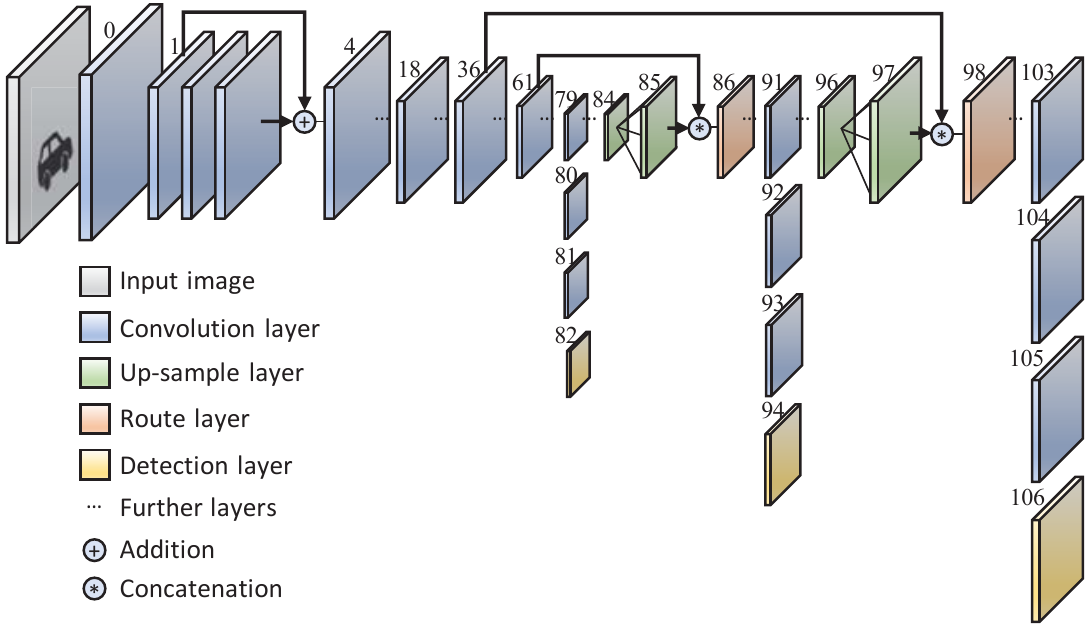}
			\caption{}
			\label{fig:a}
		\end{subfigure}
		\begin{subfigure}[t]{0.35\textwidth}
			\centering
			\includegraphics[scale=0.9]{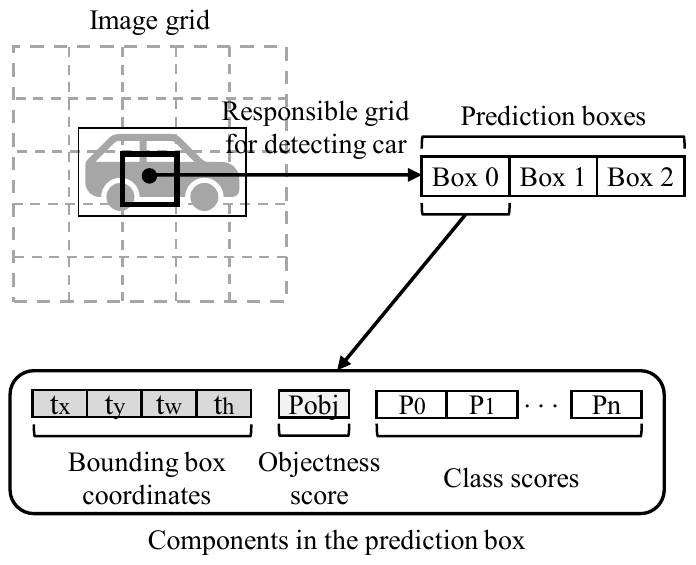}
			\caption{}
			\label{fig:b}
		\end{subfigure}\enspace
		\caption{(a) Network architecture of YOLOv3 and (b) attributes of its prediction feature map.}
		\label{fig:convent_out}
	\end{figure*}
	
	To overcome the problems of previous object detection studies, this paper proposes a novel object detection algorithm suitable for autonomous driving based on YOLOv3~\cite{redmon2018yolov3}. YOLOv3 can detect multiple objects with a single inference, and its detection speed is therefore extremely fast; in addition, by applying a multi-stage detection method, it can complement the low accuracy of YOLO~\cite{redmon2016you} and YOLOv2~\cite{redmon2017yolo9000}. Based on these advantages, YOLOv3 is suitable for autonomous driving applications, but generally achieves a lower accuracy than a two-stage method. It is therefore essential to improve the accuracy while maintaining a real-time object detection capability. To achieve this goal, the present paper proposes a method for improving the detection accuracy by modeling the bbox coordinates of YOLOv3, which only outputs deterministic values, as the Gaussian parameters (\textit{i.e.}, the mean and variance), and redesigning the loss function of bbox. Through this Gaussian modeling, a localization uncertainty for a bbox regression task in YOLOv3 can be estimated. Furthermore, to further improve the detection accuracy, a method for reducing the FP and increasing the TP by utilizing the predicted localization uncertainty of bbox during the detection process is proposed. This study is therefore the first attempt to model the localization uncertainty in YOLOv3 and to utilize this factor in a practical manner. As a result, the proposed Gaussian YOLOv3 can cope with mislocalizations in autonomous driving applications. In addition, because the proposed method is modeled only in bbox of the YOLOv3 detection layer (\textit{i.e.}, the output layer), the additional computation cost is negligible, and the proposed algorithm consequently maintains the real-time detection speed of over 42 fps with an input resolution of 512 $\times$ 512 despite the significant improvements in performance. Compared to the baseline algorithm (\textit{i.e.}, YOLOv3), the proposed Gaussian YOLOv3 improves the mAP by 3.09 and 3.5 on the KITTI~\cite{geiger2012we} and BDD~\cite{yu2018bdd100k} datasets, respectively. In addition, the proposed algorithm reduces the FP by 41.40\% and 40.62\%, respectively, and increases the TP by 7.26\% and 4.3\%, respectively, on the KITTI and BDD datasets. As a result, in terms of the trade-off between accuracy and detection speed, the proposed algorithm is suitable for autonomous driving because it significantly improves the detection accuracy and addresses the mislocalization problem while supporting a real-time operation.
	
	\section{Background}
	Instead of the region proposal method used in two-stage detectors, YOLO~\cite{redmon2016you} detects objects by dividing an image into grid units. The feature map of the YOLO output layer is designed to output bbox coordinates, the objectness score, and the class scores, and thus YOLO enables the detection of multiple objects with a single inference. Therefore, the detection speed is much faster than that of conventional methods. However, owing to the processing of the grid unit, localization errors are large and the detection accuracy is low, and thus it is unsuitable for autonomous driving applications. To address these problems, YOLOv2~\cite{redmon2017yolo9000} has been proposed. YOLOv2 improves the detection accuracy compared to YOLO by using batch normalization for the convolution layer, and applying an anchor box, multi-scale training, and fine-grained features. However, the detection accuracy is still low for small or dense objects. Therefore, YOLOv2 is unsuitable for autonomous driving applications, where a high accuracy is required for dense road objects and small objects such as traffic signs and lights.
	
	To overcome the disadvantages of YOLOv2, YOLOv3~\cite{redmon2018yolov3} has been proposed. YOLOv3 consists of convolution layers, as shown in Figure~\ref{fig:a}, and is constructed of a deep network for an improved accuracy. YOLOv3 applies a residual skip connection to solve the vanishing gradient problem of deep networks and uses an up-sampling and concatenation method that preserves fine-grained features for small object detection. The most prominent feature is the detection at three different scales in a similar manner as used in a feature pyramid network~\cite{lin2017feature}. This allows YOLOv3 to detect objects with various sizes. In more detail, when an image of three channels of R, G, and B is input into the YOLOv3 network, as shown in Figure~\ref{fig:a}, information on the object detection (\textit{i.e.}, bbox coordinates, objectness score, and class scores) is output from three detection layers. The predicted results of the three detection layers are combined and processed using non-maximum suppression. After that, the final detection results are determined. Because YOLOv3 is a fully convolutional network consisting only of small-sized convolution filers of 1 $\times$ 1 and 3 $\times$ 3 like YOLOv2~\cite{redmon2017yolo9000}, the detection speed is as fast as YOLO~\cite{redmon2016you} and YOLOv2~\cite{redmon2017yolo9000}. Therefore, in terms of the trade-off between accuracy and speed, YOLOv3 is suitable for autonomous driving applications and is widely used in autonomous driving research~\cite{corovic2018real}. However, in general, it still has a lower accuracy than a two-stage detector using a region proposal stage. To compensate for this drawback, as taking advantage of the smaller complexity of YOLOv3 than that of a two-stage detector, a more efficient detector for an autonomous driving application can be designed by applying the additional method for improving accuracy to YOLOv3~\cite{redmon2018yolov3}. The Gaussian modeling and loss function reconstruction of YOLOv3 proposed in this paper can improve the accuracy by reducing the influence of noisy data during training and predict the localization uncertainty. In addition, the detection accuracy can be further enhanced by using this predicted localization uncertainty. A detailed description of the above aspects is provided in Section 3.
	
	\section{Gaussian YOLOv3}
	\subsection{Gaussian modeling}
	As shown in Figure~\ref{fig:b}, the prediction feature map of YOLOv3~\cite{redmon2018yolov3} has three prediction boxes per grid, where each prediction box consists of bbox coordinates (\textit{i.e.}, $t_{x}$, $t_{y}$, $t_{w}$, and $t_{h}$), the objectness score, and class scores. YOLOv3 outputs the objectness (\textit{i.e.}, whether an object is present or not in the bbox) and class (\textit{i.e.}, the category of the object), as a score of between zero and one. An object is then detected based on the product of these two values. Unlike the objectness and class information, bbox coordinates are output as deterministic coordinate values instead of a score, and thus the confidence of the detected bbox is unknown. Moreover, the objectness score does not reflect the reliability of the bbox well. It therefore does not know how uncertain the result of bbox is. In contrast, the uncertainty of bbox, which is predicted by the proposed method, serves as the bbox score, and can thus be used as an indicator of how uncertain the bbox is. The results for this are described in Section 4.1.
	
	In YOLOv3, bbox regression is to extract the bbox center information (\textit{i.e.}, $t_{x}$ and $t_{y}$) and bbox size information (\textit{i.e.}, $t_{w}$ and $t_{h}$). Because there is only one correct answer (\textit{i.e.}, the GT) for the bbox of an object, complex modeling is not required for predicting the localization uncertainty. In other words, the uncertainty of bbox can be modeled using each single Gaussian model of $t_{x}$, $t_{y}$, $t_{w}$, and $t_{h}$. A single Gaussian model of output $y$ for a given test input $x$ whose output consists of Gaussian parameters is as follows:
	\begin{equation}
	p(y|x)= N(y;\mu(x),\Sigma(x)),
	\end{equation}
	where $\mu(x)$ and $\Sigma(x)$ are the mean and variance functions, respectively.
	
\begin{figure}[t]
		\centering
		\includegraphics[scale=0.9]{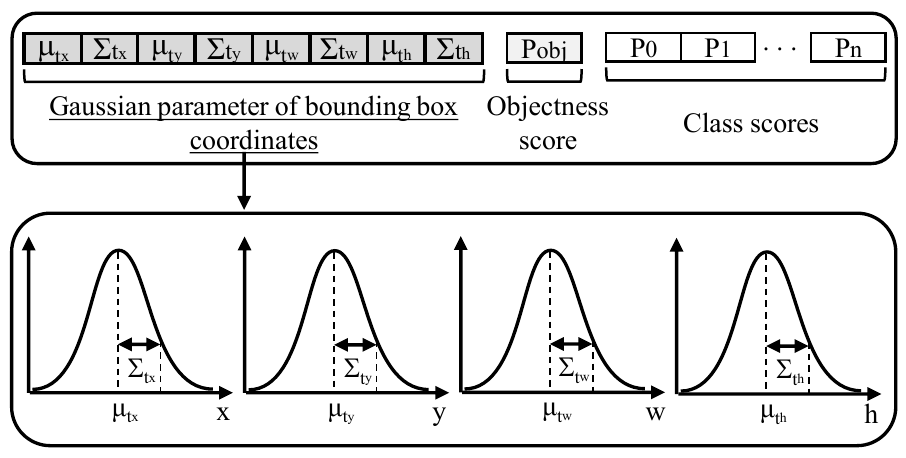}
		\caption{Components in the prediction box of proposed algorithm.}
		\label{fig:proposed_out}
	\end{figure}

	To predict the uncertainty of bbox, each of the bbox coordinates in the prediction feature map is modeled as the mean ($\mu$) and variance ($\Sigma$), as shown in Figure~\ref{fig:proposed_out}. The outputs of bbox are $\hat{\mu}_{t_{x}}$, $\hat{\Sigma}_{t_{x}}$, $\hat{\mu}_{t_{y}}$, $\hat{\Sigma}_{t_{y}}$, $\hat{\mu}_{t_{w}}$, $\hat{\Sigma}_{t_{w}}$, $\hat{\mu}_{t_{h}}$, and $\hat{\Sigma}_{t_{h}}$. Considering the structure of the detection layer in YOLOv3, the Gaussian parameters for $t_{x}$, $t_{y}$, $t_{w}$, and $t_{h}$ are preprocessed as follows:
	\begin{equation}
	\mu_{t_{x}}=\sigma(\hat{\mu}_{t_{x}}),~\mu_{t_{y}}=\sigma(\hat{\mu}_{t_{y}}),~\mu_{t_{w}}=\hat{\mu}_{t_{w}},~\mu_{t_{h}}=\hat{\mu}_{t_{h}} 
	\end{equation}
	\begin{equation}
	\begin{aligned}
	\Sigma_{t_{x}}=\sigma(\hat{\Sigma}_{t_{x}}),~\Sigma_{t_{y}}=\sigma(\hat{\Sigma}_{t_{y}}),
	\\
	~\Sigma_{t_{w}}=\sigma(\hat{\Sigma}_{t_{w}}),~\Sigma_{t_{h}}=\sigma(\hat{\Sigma}_{t_{h}}) 
	\end{aligned}
	\end{equation}
	\begin{equation}
	\sigma(x)= \frac{1}{(1+exp(-x))}. 
	\end{equation}
	The mean value of each coordinate in the detection layer is the predicted coordinate of bbox, and each variance represents the uncertainty of each coordinate. $\mu_{t_{x}}$ and $\mu_{t_{y}}$ in (2) must represent the center coordinates of bbox inside the grid, which are thus processed as values between zero and one with the sigmoid function in (4). The variances of each coordinate in (3) are also processed as values between zero and one with a sigmoid function. In YOLOv3, the $width$ and $height$ information of bbox are processed through $t_w$, $t_h$, bbox prior, and exponential functions~\cite{redmon2018yolov3}. In other words, $\mu_{t_{w}}$ and $\mu_{t_{h}}$ in (2), which indicate the $t_w$ and $t_h$ of YOLOv3, are not processed as sigmoid functions because they can have both negative and positive values.
	
	Single Gaussian modeling for predicting the uncertainty of bbox only applies to the bbox coordinates of the YOLOv3 detection layer shown in Figure~\ref{fig:a}. Therefore, the overall computational complexity of the algorithm does not increase significantly. In a 512 $\times$ 512 input resolution and ten classes, YOLOv3 requires 99 $\times$ $10^9$ FLOPs; however, after a single Gaussian modeling for bbox, 99.04 $\times$ $10^9$ FLOPs are required. Thus, the penalty for the detection speed is extremely low because the computation cost increases only by 0.04\% as compared with before the modeling. The related results are shown in Section 4.
	\subsection{Reconstruction of loss function}
	For training, YOLOv3~\cite{redmon2018yolov3} uses the sum of the squared error loss for bbox, and the binary cross-entropy loss for the objectness and class. Because the bbox coordinates are output as Gaussian parameters through Gaussian modeling, the loss function of bbox is redesigned as a negative log likelihood (NLL) loss, whereas the loss function for objectness and class is not changed. The loss function redesigned for bbox is as follows:
	\begin{equation}
	\begin{aligned}
	L_x=-\sum_{i=1}^W\sum_{j=1}^H\sum_{k=1}^K\gamma_{ijk}
	log(N(x^G_{ijk}|\mu_{t_{x}}(x_{ijk}),
	\\
	\Sigma_{t_{x}}(x_{ijk}))+\varepsilon),
	\end{aligned}
	\end{equation}
	where $L_x$ is the NLL loss of $t_x$ coordinate and the others (\textit{i.e.}, $L_y$, $L_w$, and $L_h$) are the same as $L_x$ except for each parameter. $W$ and $H$ are the number of grids of each $width$ and $height$, respectively, and $K$ is the number of anchors. Moreover, $\mu_{t_{x}}(x_{ijk})$ denotes the $t_x$ coordinates, which is the output of the detection layer of the proposed algorithm, at the \textit{k}-th anchor in the (\textit{i}, \textit{j}) grid. In addition, $\Sigma_{t_{x}}(x_{ijk})$ is also the output of the detection layer, indicating the uncertainty of $t_x$ coordinate, and $x^G_{ijk}$ is the GT of $t_x$ coordinate. The GT of bbox is then computed as follows:
	\begin{equation}
	x^G_{ijk}=x^G\times W-i,\ y^G_{ijk}=y^G \times H-j 
	\end{equation}
	\begin{equation}
	w^G_{ijk}=log(\frac{w^G \times IW}{A^w_k}),\ h^G_{ijk}=log(\frac{h^G \times IH}{A^h_k}),
	\end{equation}
	where $x^G$, $y^G$, $w^G$, and $h^G$ are the ratios of a GT bbox in an image, $IW$ and $IH$ are the $width$ and $height$ of the resized image, and $A^w_k$ and $A^h_k$ denote the $width$ and $height$ of the \textit{k}-th anchor box prior, respectively. In YOLOv3, centroid of bbox is calculated in grid units, and size of bbox is calculated based on an anchor box, and thus the GT is processed accordingly for training.
	\begin{equation}
	\gamma_{ijk} = \frac{\omega_{scale} \times \delta^{obj}_{ijk}}{2}
	\end{equation}
	\begin{equation}
	\omega_{scale}=2-w^G \times h^G.
	\end{equation}
	$\omega_{scale}$ in (8) is calculated based on the $width$ and $height$ ratios of the GT bbox in an image, as shown in (9). It provides different weights according to the object size during training. In addition, $\delta^{obj}_{ijk}$ in (8) is a parameter applied to include in the loss only if there is an anchor that is most suitable in the current object among the predefined anchors. This parameter is assigned as a value of one when the intersection over union (IOU) of the GT and the \textit{k}-th anchor box in the (\textit{i}, \textit{j}) grid are the largest, and is assigned as a value of zero if there is no appropriate GT. For a numerical stability of the logarithmic function, $\varepsilon$ is assigned a value of $10^{-9}$.
	
	Because YOLOv3 uses the sum of the squared error loss for bbox, it is unable to cope with noisy data during training. However, the redesigned loss function of bbox can provide a penalty to the loss through the uncertainty for inconsistent data during training. That is, the model can be learned by concentrating on consistent data. Therefore, the redesigned loss function of bbox makes the model more robust to noisy data~\cite{kendall2017uncertainties}. Through this loss attenuation~\cite{kendall2017uncertainties}, it is possible to improve the accuracy of the algorithm.
	
	\subsection{Utilization of localization uncertainty}
	The proposed Gaussian YOLOv3 can obtain the uncertainty of bbox for every detection object in an image. Because it is not an uncertainty for the entire image, it is possible to apply uncertainty to each detection result. YOLOv3 considers only the objectness score and class scores during object detection, and cannot consider the bbox score during the detection process because the score information for the bbox coordinates is unknown. However, Gaussian YOLOv3 can output the localization uncertainty, which is the score of bbox. Therefore, localization uncertainty can be considered along with the objectness score and class scores during the detection process. The proposed algorithm applies localization uncertainty to the detection criteria of YOLOv3 such that bbox with high uncertainty among the predicted results is filtered through the detection process. In this way, predictions with high confidence of objectness, class, and bbox are finally selected. Thus, Gaussian YOLOv3 can reduce the FP and increase the TP, which results in improving the detection accuracy. The proposed detection criterion considering the localization uncertainty is as follows:
	\begin{equation}
	Cr.=\sigma(Object)\times\sigma(Class_i)\times(1-Uncertainty_{aver}).
	\end{equation}
	$Cr.$ in (10) indicates the detection criterion for Gaussian YOLOv3, $\sigma(Object)$ is the objectness score, and $\sigma(Class_i)$ is the score of the \textit{i}-th class. In addition, $Uncertainty_{aver}$, which is localization uncertainty, indicates the average of the uncertainties of the predicted bbox coordinates. Localization uncertainty has a value between zero and one, such as the objectness score and class scores, and the higher the localization uncertainty, the lower the confidence of the predicted bbox. The results of the proposed Gaussian YOLOv3 are described in Section 4.
	
	\section{Experimental Results}
	In the experiment, the KITTI dataset~\cite{geiger2012we}, which is commonly used in autonomous driving research, and the BDD dataset~\cite{yu2018bdd100k}, which is the latest published autonomous driving dataset, are used. The KITTI dataset consists of three classes: car, cyclist, and pedestrian, and consists of 7,481 images for training and 7,518 images for testing. Because there is no GT for testing, the training and validation sets are made by randomly splitting the training set in half~\cite{wu2017squeezedet}. The BDD dataset consists of ten classes: bike, bus, car, motor, person, rider, traffic light, traffic sign, train, and truck. The ratio of training, validation, and test set is 7:1:2. In this paper, a test set is utilized for the performance evaluation. In general, the IOU threshold (TH) of the KITTI dataset is set to 0.7 for cars and 0.5 for cyclists and pedestrians~\cite{geiger2012we}, whereas the IOU TH of the BDD dataset is 0.75 for all classes~\cite{yu2018bdd100k}. In both YOLOv3 and Gaussian YOLOv3 training, the batch size is 64 and the learning rate is 0.0001. The anchor size is extracted using k-means clustering for each training set of KITTI and BDD. The anchors used in the training and evaluation are shown in Table~\ref{table_cluster}. Other studies are trained using the default settings in the official code of each algorithm. The experiment is conducted on an NVIDIA GTX 1080 Ti with CUDA 8.0 and cuDNN v7.
	
	\subsection{Validation in utilizing localization uncertainty}
\begin{figure}[t]
		\centering
		\includegraphics[scale=0.86]{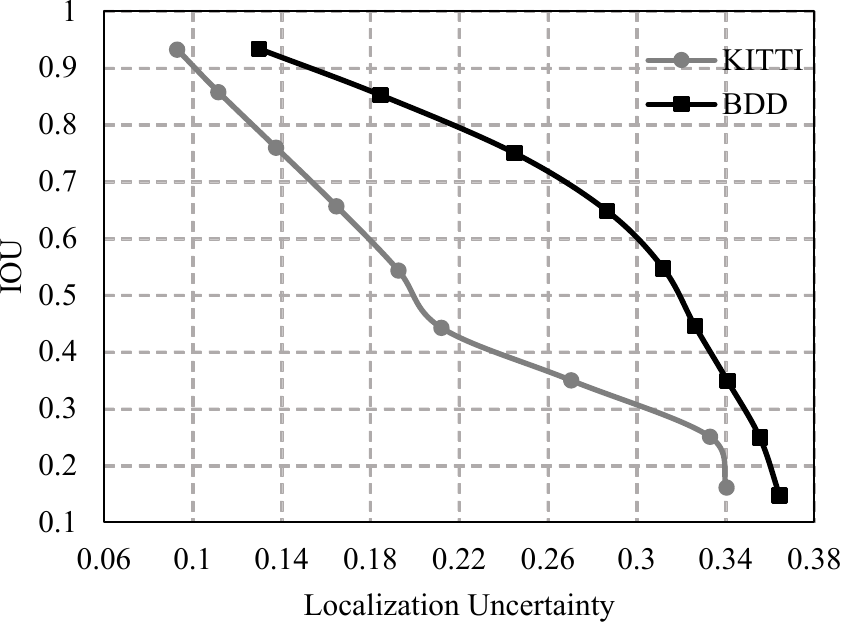}
		\caption{IOU versus localization uncertainty on KITTI and BDD validation sets.}
		\label{fig:iou_uc}
	\end{figure}

	Figure~\ref{fig:iou_uc} shows the relationship between the IOU and localization uncertainty of bbox for the KITTI and BDD validation sets. These results are plotted for cars, which is the dominant class for all data, and the localization uncertainty is predicted using the proposed algorithm. To show a typical tendency, the IOU is divided increments of 0.1, and the average value of the IOU and the average value of the localization uncertainty are calculated for each range and used as a representative value. As shown in Figure~\ref{fig:iou_uc}, the IOU value tends to increase as the localization uncertainty decreases in both datasets. A larger IOU indicates that the coordinates of the predicted bbox are closer to those of the GT. Based on these results, the localization uncertainty of the proposed algorithm effectively represents the confidence of the predicted bbox. It is therefore possible to cope with mislocalizations and improve the accuracy by utilizing the localization uncertainty predicted by the proposed algorithms.
	\begin{table}[t]
		\centering
		\resizebox{0.41\textwidth}{!}
		{%
			\begin{tabular}{cccc}
				\hline\hline
				& Anchor 0 & Anchor 1  & Anchor 2  \\ \hline
				KITTI training set     &          &           &           \\
				First detection layer  & (49,240) & (82,170)  & (118,206) \\
				Second detection layer & (45,76)  & (27,172)  & (67,116)  \\
				Third detection layer  & (13,30)  & (23,53)   & (17,102)   \\ \hline
				BDD training set       &          &           &           \\
				First detection layer  & (73,175) & (141,178) & (144,291) \\
				Second detection layer & (32,97)  & (57,64)   & (92,109)  \\
				Third detection layer  & (7,10)   & (14,24)   & (27,43)   \\ \hline\hline
			\end{tabular}
		}
		\caption{Results of anchor boxes of training sets.}
		\label{table_cluster}
	\end{table}
	\subsection{Performance evaluation of Gaussian YOLOv3}
	\begin{table*}[!t]
		\centering
		\resizebox{0.94\textwidth}{!}
		{
			\begin{tabular}{ccccccccccccc}
				\hline\hline
				\multirow{3}{*}{Detection algorithm} & \multicolumn{9}{c}{Average precision (\%)}                                             & \multirow{3}{*}{mAP (\%)} & \multirow{3}{*}{FPS} & \multirow{3}{*}{Input size} \\
				& \multicolumn{3}{c}{Car} & \multicolumn{3}{c}{Pedestrian} & \multicolumn{3}{c}{Cyclist} &                           &                      &                             \\
				& E      & M      & H     & E        & M        & H        & E       & M       & H       &                           &                      &                             \\ \hline
				MS-CNN~\cite{cai2016unified}                               & 92.54  & 90.49  & 79.23 & 87.46    & 81.34    & 72.49    & 90.13   & 87.59   & 81.11   & 84.71                     & 8.13                 & 1920$\times$576                    \\
				SINet~\cite{hu2019sinet}                                & 99.11  & 90.59  & 79.77 & 88.09    & 79.22    & 70.30    & 94.41   & 86.61   & 80.68   & 85.42                     & 23.98                & 1920$\times$576                    \\
				SSD~\cite{liu2016ssd}                                  & 88.37  & 87.84  & 79.15 & 50.33    & 48.87    & 44.97    & 48.00   & 52.51   & 51.52   & 61.29                     & 28.93                & 512$\times$512                     \\
				RefineDet~\cite{zhang2018single}                            & 98.96  & 90.44  & 88.82 & 84.40    & 77.44    & 73.52    & 86.33   & 80.22   & 79.15   & 84.36                     & 27.81                & 512$\times$512                     \\
				CFENet~\cite{zhao2018cfenet}                               & 90.33  & 90.22  & 84.85 & -        & -        & -        & -       & -       & -       & -                         & 0.25                 & -                           \\
				RFBNet~\cite{liu2018receptive}                               & 87.41  & 88.35  & 83.41 & 65.85    & 61.30    & 57.71    & 74.46   & 72.73   & 69.75   & 73.44                     & 39.20                & 512$\times$512                     \\
				YOLOv3~\cite{redmon2018yolov3}                               & 85.68  & 76.89  & 75.89 & 83.51    & 78.37    & 75.16    & 88.94   & 80.64   & 79.62   & 80.52                     & 43.57                & 512$\times$512                     \\ \hline
				Gaussian YOLOv3                      & 90.61  & 90.20  & 81.19 & 87.84    & 79.57    & 72.30    & 89.31   & 81.30   & 80.20   & 83.61                     & 43.13                & 512$\times$512                     \\
				Gaussian YOLOv3                      & 98.74  & 90.48  & 89.47 & 87.85    & 79.96    & 76.81    & 90.08   & 86.59   & 81.09   & 86.79                     & 24.91                & 704$\times$704                     \\ \hline\hline
			\end{tabular}
		}
		\caption{Performance comparison using KITTI validation set. E, M, and H refer to easy, moderate, and hard, respectively.}
		\label{table_kitti_comparosion}
	\end{table*}
	
	\begin{table}[t]
		\centering
		\resizebox{0.41\textwidth}{!}
		{
			\begin{tabular}{cccc}
				\hline\hline
				Detection algorithm & mAP (\%) & FPS  & Input size \\ \hline
				MS-CNN~\cite{cai2016unified}              & 5.7      & 6.0  & 1920$\times$576   \\
				SINet~\cite{hu2019sinet}               & 9.0        & 18.2 & 1920$\times$576   \\
				SSD~\cite{liu2016ssd}                 & 14.1     & 23.1 & 512$\times$512    \\
				RefineDet~\cite{zhang2018single}           & 17.4     & 22.3 & 512$\times$512    \\
				CFENet~\cite{zhao2018cfenet}              & 19.1     & 21.0 & 512$\times$512    \\
				RFBNet~\cite{liu2018receptive}              & 14.5        & 39.0    & 512$\times$512    \\
				YOLOv3~\cite{redmon2018yolov3}              & 14.9     & 42.9    & 512$\times$512    \\ \hline
				Gaussian YOLOv3     & 18.4     & 42.5    & 512$\times$512    \\
				Gaussian YOLOv3     & 20.8     & 22.5    & 736$\times$736    \\ \hline\hline
			\end{tabular}
		}
		\caption{Performance comparison using BDD test set.}
		\label{table_bdd_comparosion}
	\end{table}
	To demonstrate the superiority of the proposed algorithm, its performance (\textit{i.e.}, accuracy and detection speed) is compared with that of other studies~\cite{cai2016unified,hu2019sinet,liu2016ssd,zhang2018single,zhao2018cfenet,liu2018receptive,redmon2018yolov3}. In the experiment on the KITTI validation set, the other studies~\cite{cai2016unified,hu2019sinet,liu2016ssd,zhang2018single,liu2018receptive,redmon2018yolov3} are trained and evaluated using the official published code of each algorithm. In the case of CFENet~\cite{zhao2018cfenet}, the result of the KITTI object detection leader board is used because the official code has not been published. In the experiment on the BDD test data, the results for the BDD test set of SSD~\cite{liu2016ssd}, CFENet~\cite{zhao2018cfenet}, and RefineDet~\cite{zhang2018single} are specified in CFENet~\cite{zhao2018cfenet}, and thus the simulation results of these studies are from~\cite{zhao2018cfenet}, whereas the remaining comparative studies~\cite{cai2016unified,hu2019sinet,liu2018receptive,redmon2018yolov3} are trained and evaluated using the official published codes because these studies have not been developed as targets for BDD datasets and therefore have not been evaluated with BDD datasets in previous studies. For a fair comparison of the one-stage detectors, the input resolution is set as in CFENet~\cite{zhao2018cfenet}. The two-stage detector uses the default resolution of each official published code. The official evaluation method of each dataset is used for an accuracy comparison, and IOU TH is set to the value mentioned before. For a comparison of the accuracy, mAP, which has been widely used in previous studies on object detection, is selected.
	
	Table~\ref{table_kitti_comparosion} shows the performance of the proposed algorithm and other methods using the KITTI validation set. The mAP of the proposed algorithm, Gaussian YOLOv3, improves by 3.09 compared to that of YOLOv3, and the detection speed is 43.13 fps, which enables real-time detection with a slight difference from YOLOv3. Gaussian YOLOv3 is 3.93 fps faster than that of RFBNet~\cite{liu2018receptive}, which has the fastest operation speed among the previous studies with the exception of YOLOv3, despite the mAP of Gaussian YOLOv3 outperforming that of RFBNet~\cite{liu2018receptive} by more than 10.17. In addition, although the mAP of Gaussian YOLOv3 with a 512 $\times$ 512 resolution is 1.81 lower than that of SINet~\cite{hu2019sinet}, which has the highest accuracy among the previous methods, it is noteworthy that the fps of the proposed method is 1.8-times better than that of SINet~\cite{hu2019sinet}. Because there is a trade-off between the accuracy and detection speed, for a fair comparison, the input resolution of the proposed algorithm is changed and evaluated considering the fps of SINet~\cite{hu2019sinet}. The experimental results show that the mAP of Gaussian YOLOv3 with a 704 $\times$ 704 resolution shown in the last row of Table~\ref{table_kitti_comparosion} is 86.79 at 24.91 fps, and consequently, Gaussian YOLOv3 outperforms SINet~\cite{hu2019sinet} in terms of the accuracy and detection speed.
	
	Table~\ref{table_bdd_comparosion} shows the performance of the proposed approach and other methods for the BDD test set. Gaussian YOLOv3 improves the mAP by 3.5 compared with YOLOv3, and the detection speed is 42.5 fps, which is almost the same as YOLOv3. In addition, Gaussian YOLOv3 is 3.5 fps faster than the RFBNet~\cite{liu2018receptive}, which has the fastest operation speed among the previous studies except for YOLOv3, despite the accuracy of Gaussian YOLOv3 outperforming that of RFBNet~\cite{liu2018receptive} by 3.9 mAP. In addition, compared to CFENet~\cite{zhao2018cfenet}, which has the highest accuracy among the previous methods, the performance of Gaussian YOLOv3 with a 736 $\times$ 736 input resolution in the last row of Table~\ref{table_bdd_comparosion} shows a better mAP of 1.7 and faster operation speed of 1.5 fps, and consequently, Gaussian YOLOv3 outperforms CFENet~\cite{zhao2018cfenet} in terms of the accuracy and detection speed.
	
	Furthermore, on the COCO dataset~\cite{lin2014microsoft}, the AP of Gaussian YOLOv3 is 36.1, which is 3.1 higher than YOLOv3. In particular, the AP$_{75}$ (\textit{i.e.}, strict metric) of Gaussian YOLOv3 is 39.0, which is 4.6 higher than that of YOLOv3. These results indicate that the proposed algorithm outperforms YOLOv3 in general dataset as well as KITTI and BDD.
	
	Based on these experimental results, because the proposed algorithm can significantly improve the accuracy with little penalty in speed compared to YOLOv3, Gaussian YOLOv3 is superior to the previous methods.
	
	\subsection{Visual and numerical evaluation of FP and TP}
\begin{figure*}[t!]
		\centering
		\includegraphics[scale=0.17]{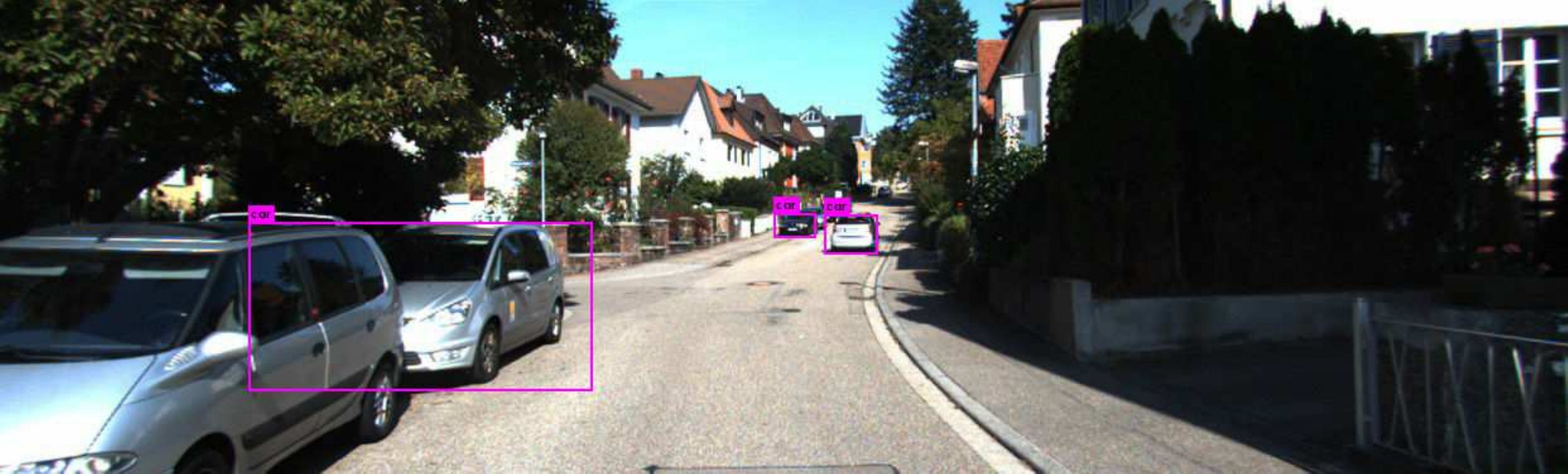}
		\includegraphics[scale=0.17]{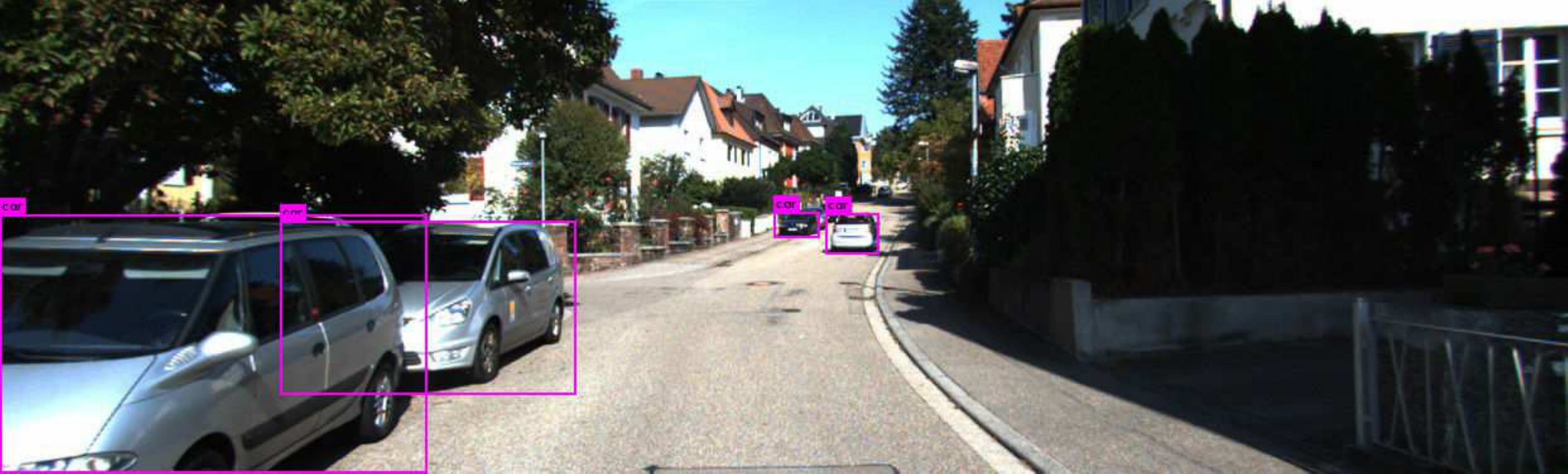}
		\includegraphics[scale=0.17]{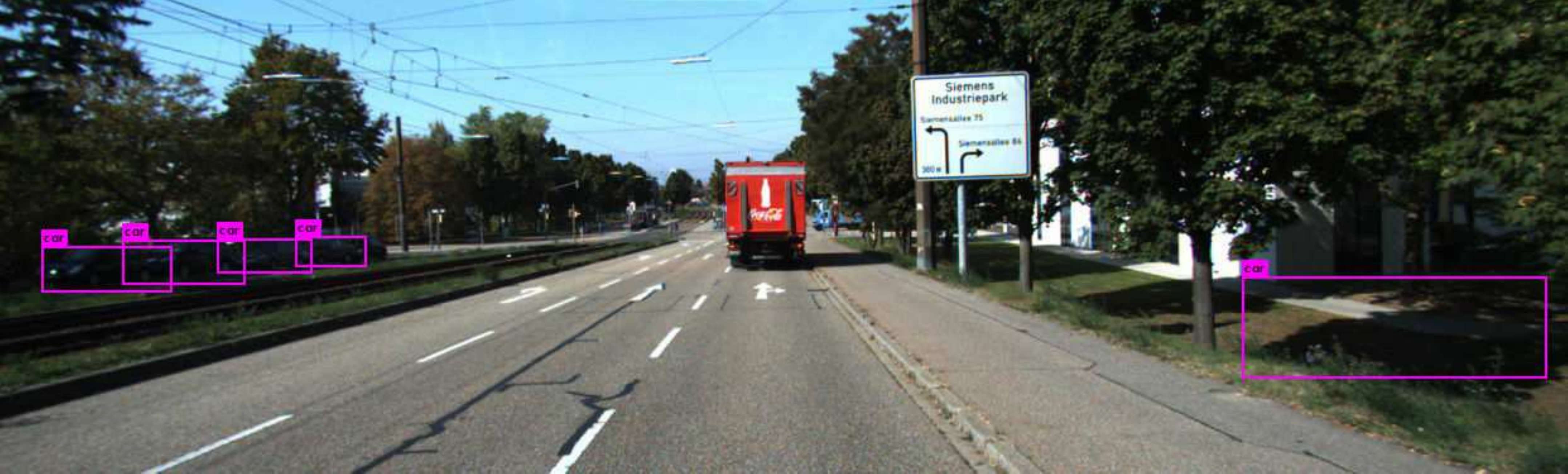}
		\includegraphics[scale=0.17]{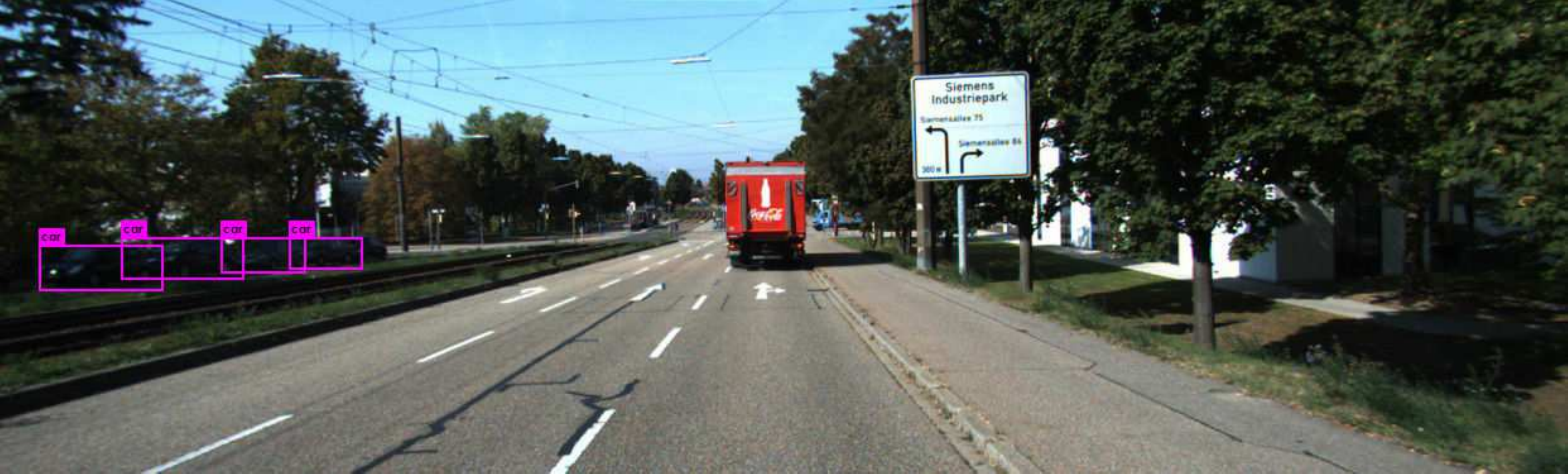}
		\includegraphics[scale=0.17]{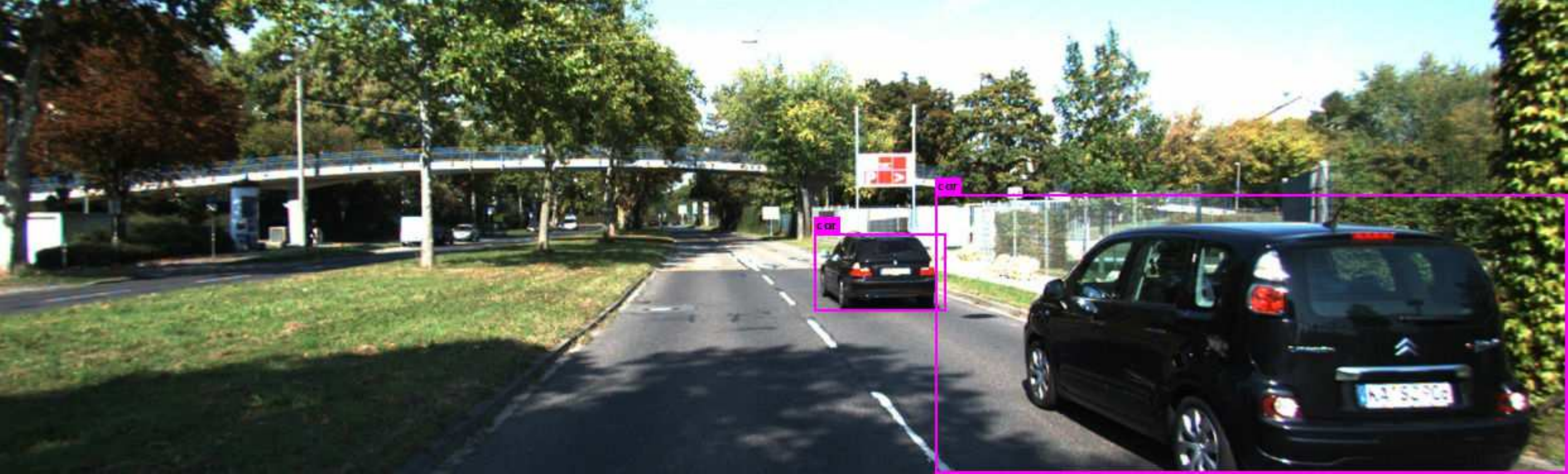}
		\includegraphics[scale=0.17]{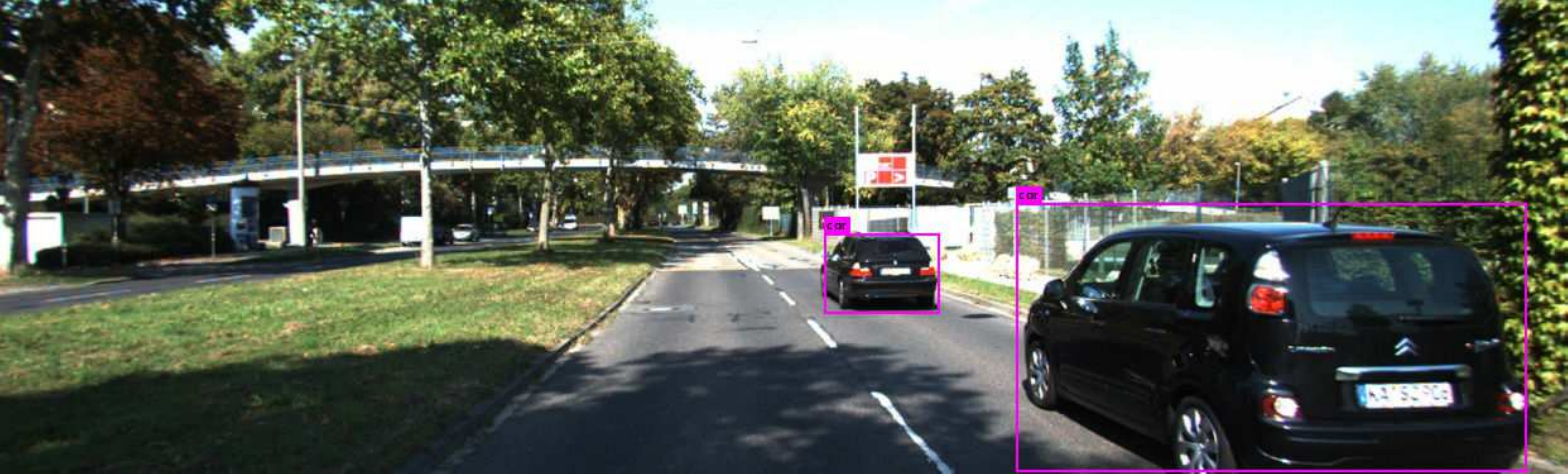}	
		\caption{Detection results of the baseline and proposed algorithms on the KITTI validation set. The first column shows the detection results of YOLOv3, whereas the second column shows the detection results of Gaussian YOLOv3.}
		\label{fig:kitti_det_result}
	\end{figure*}

\begin{figure*}[t!]
		\centering
		\includegraphics[scale=0.12]{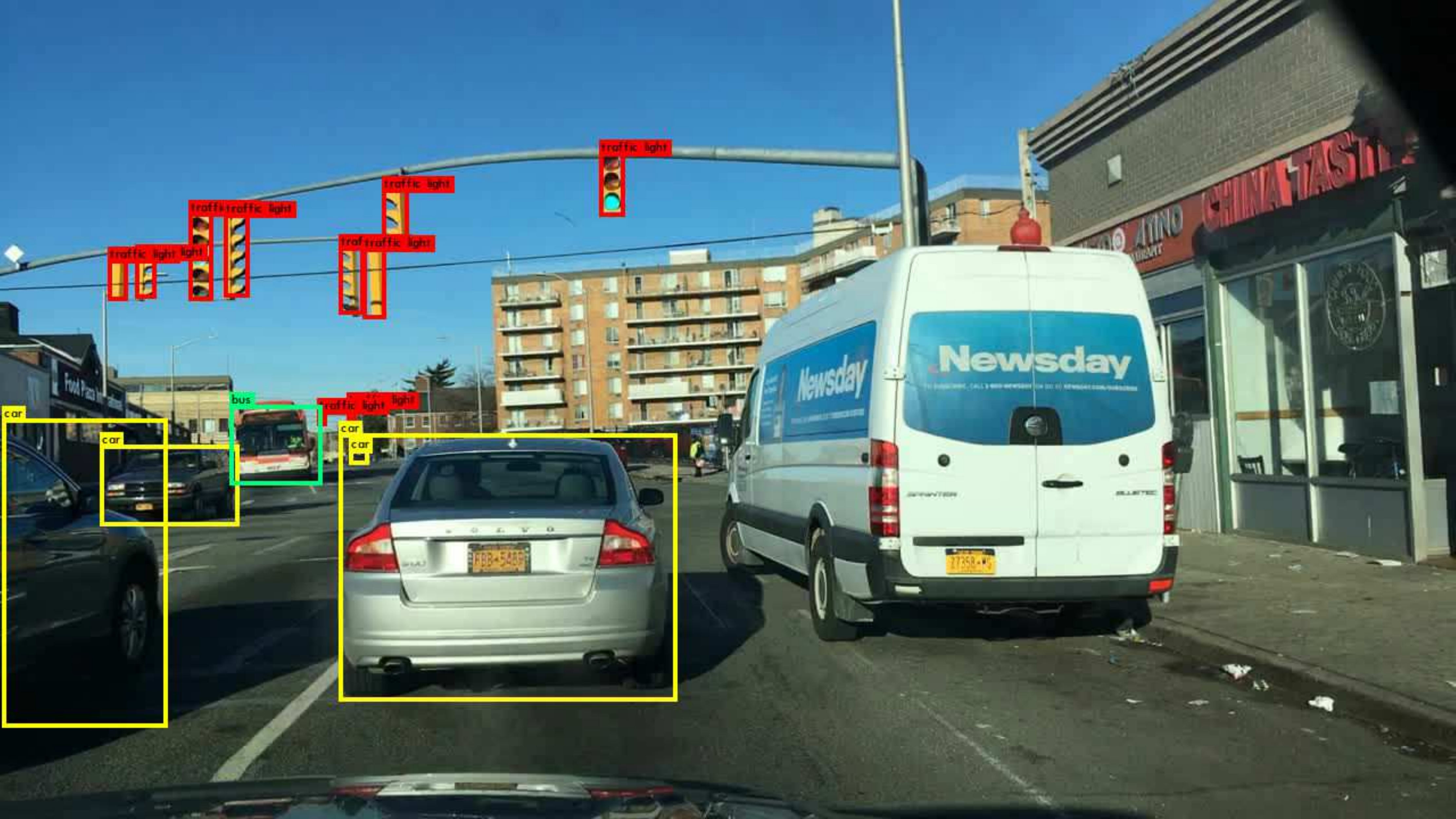}
		\includegraphics[scale=0.12]{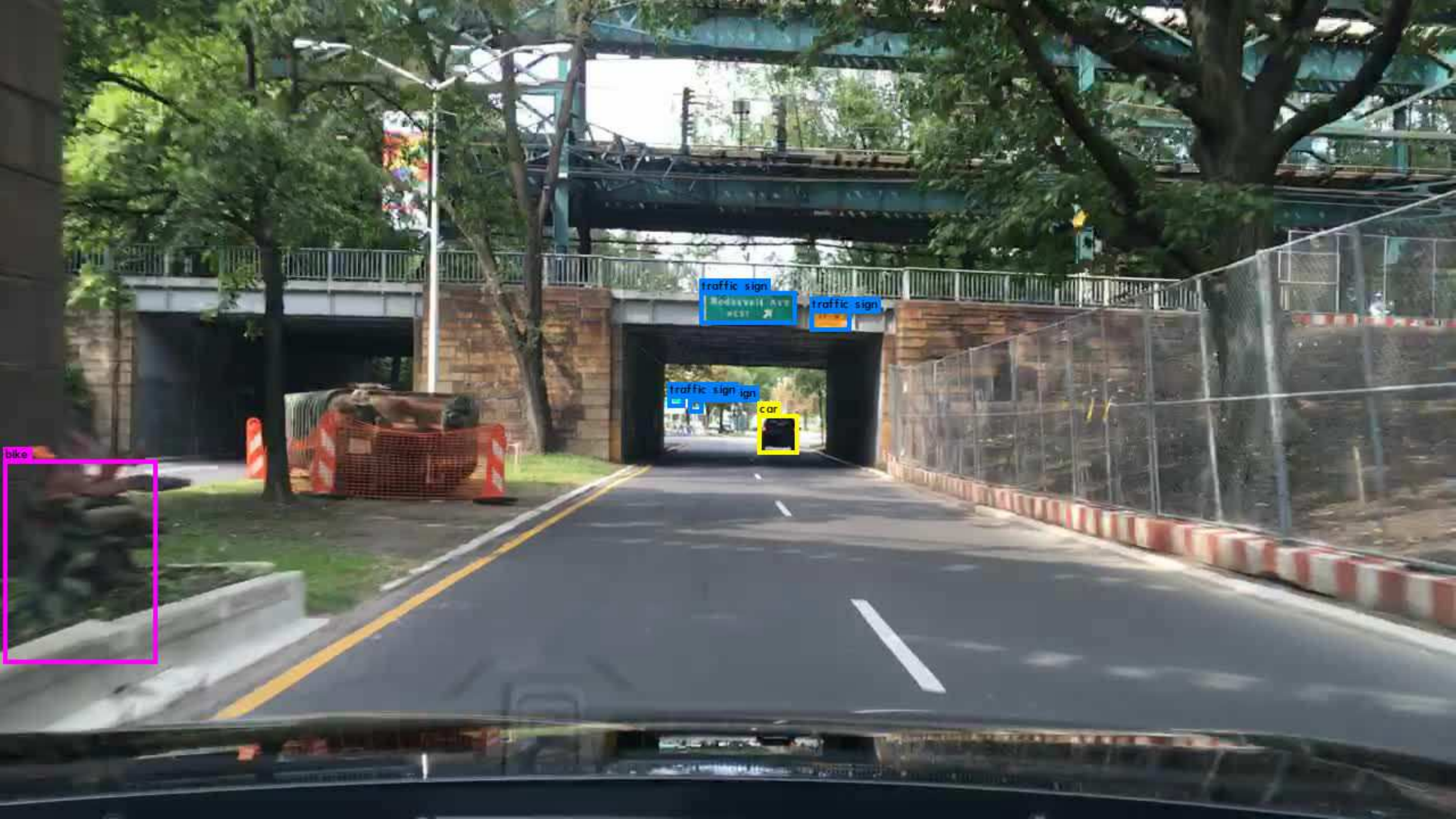}
		\includegraphics[scale=0.12]{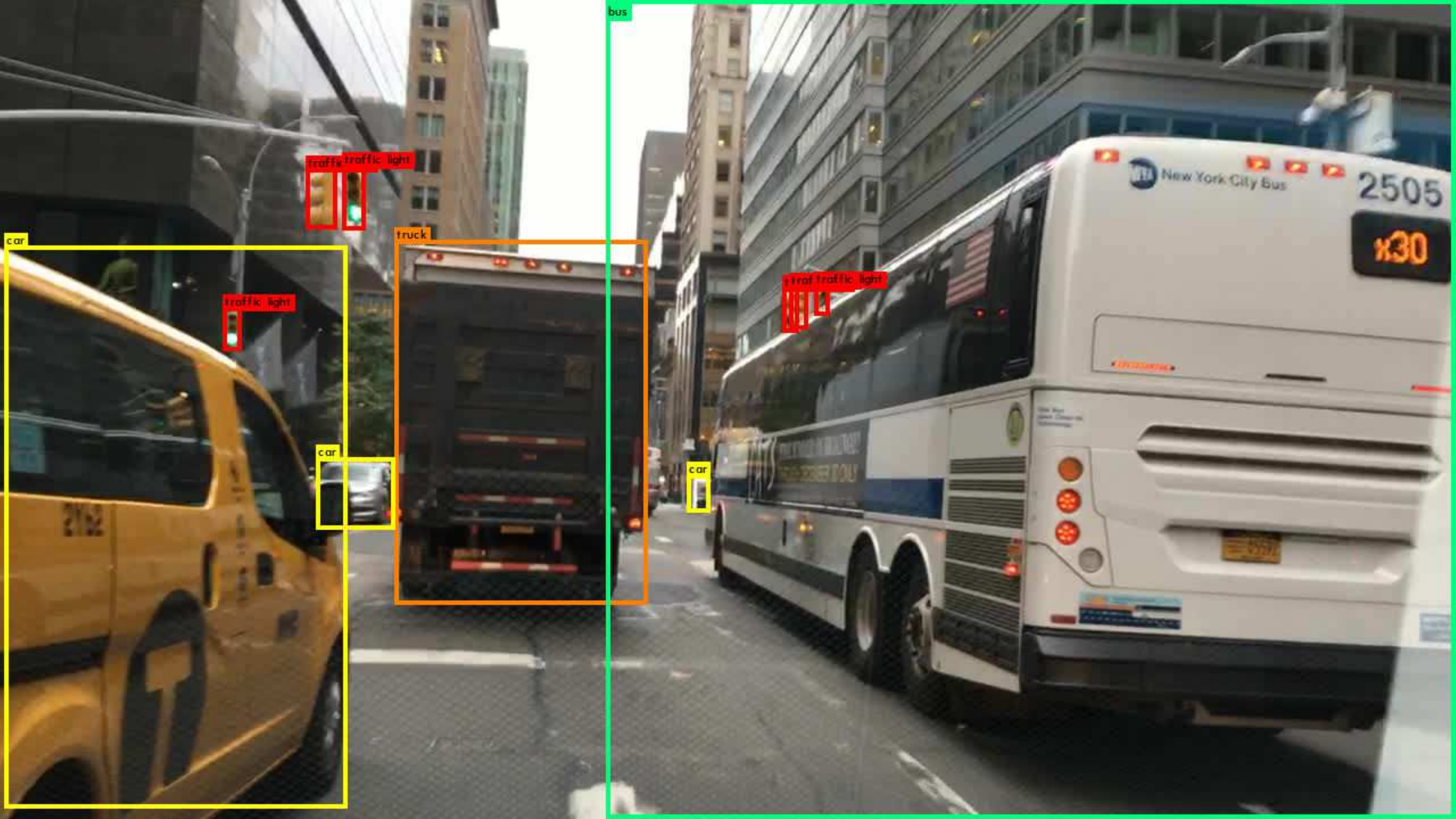}
		\includegraphics[scale=0.12]{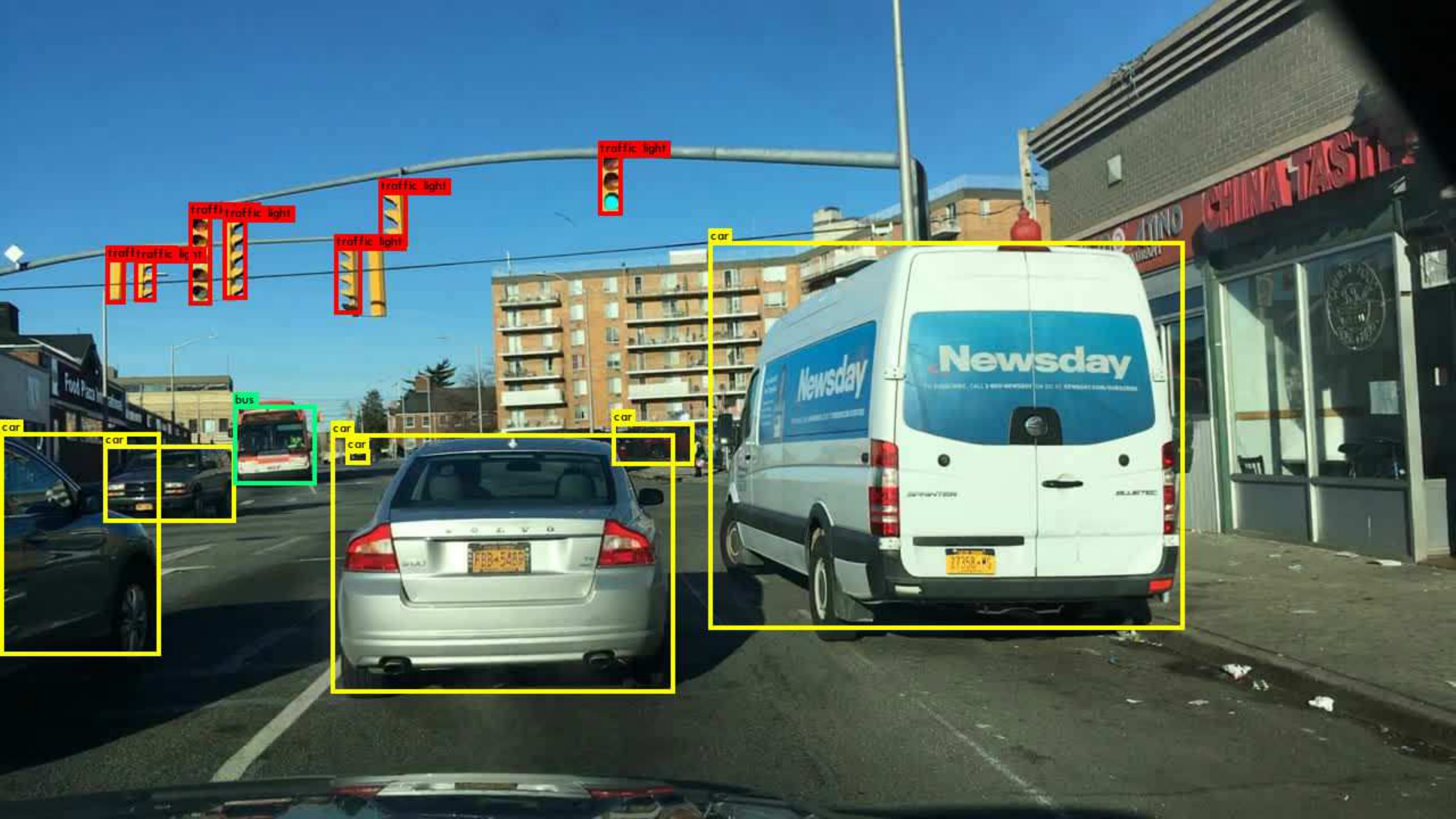}
		\includegraphics[scale=0.12]{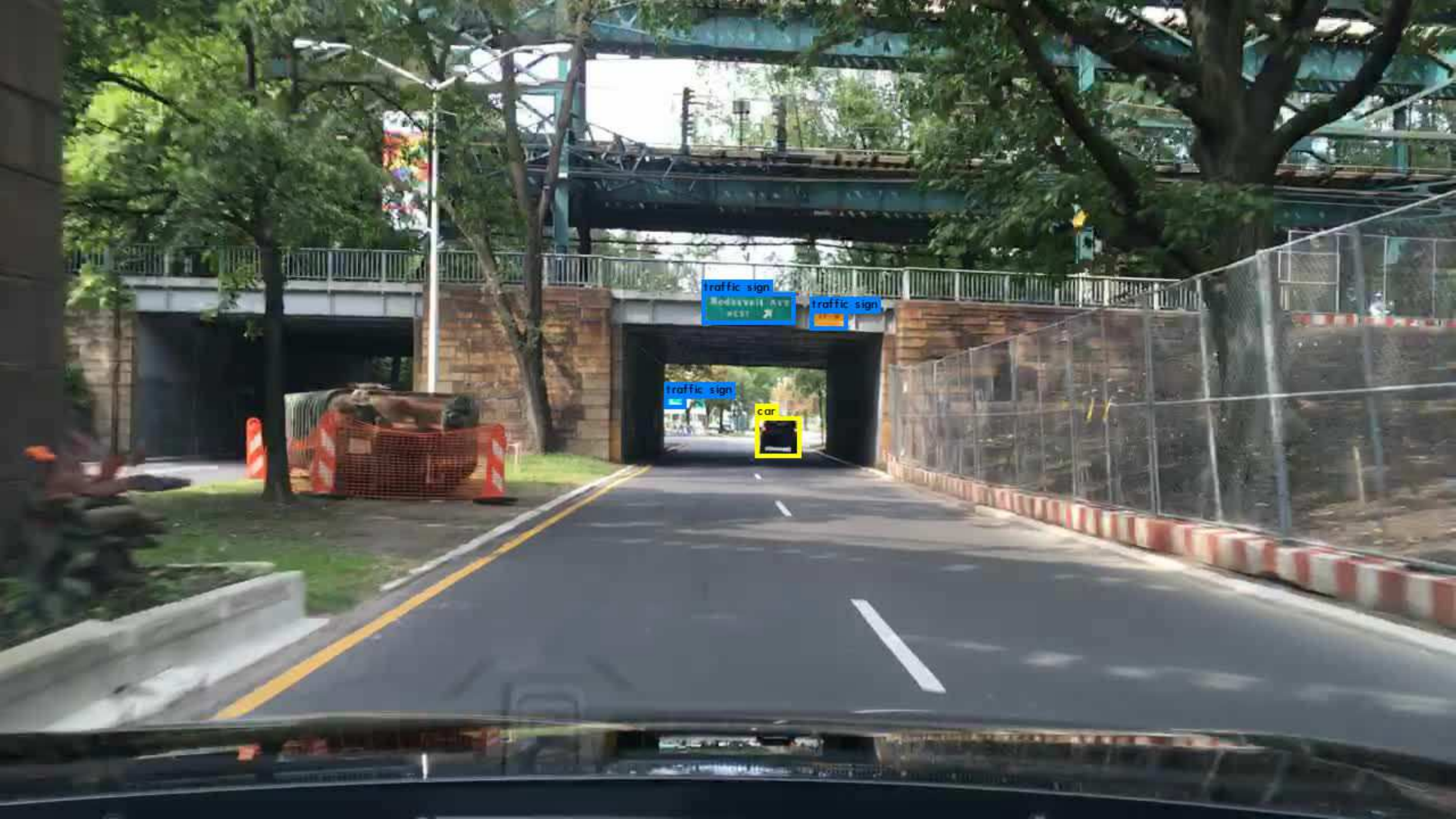}		
		\includegraphics[scale=0.12]{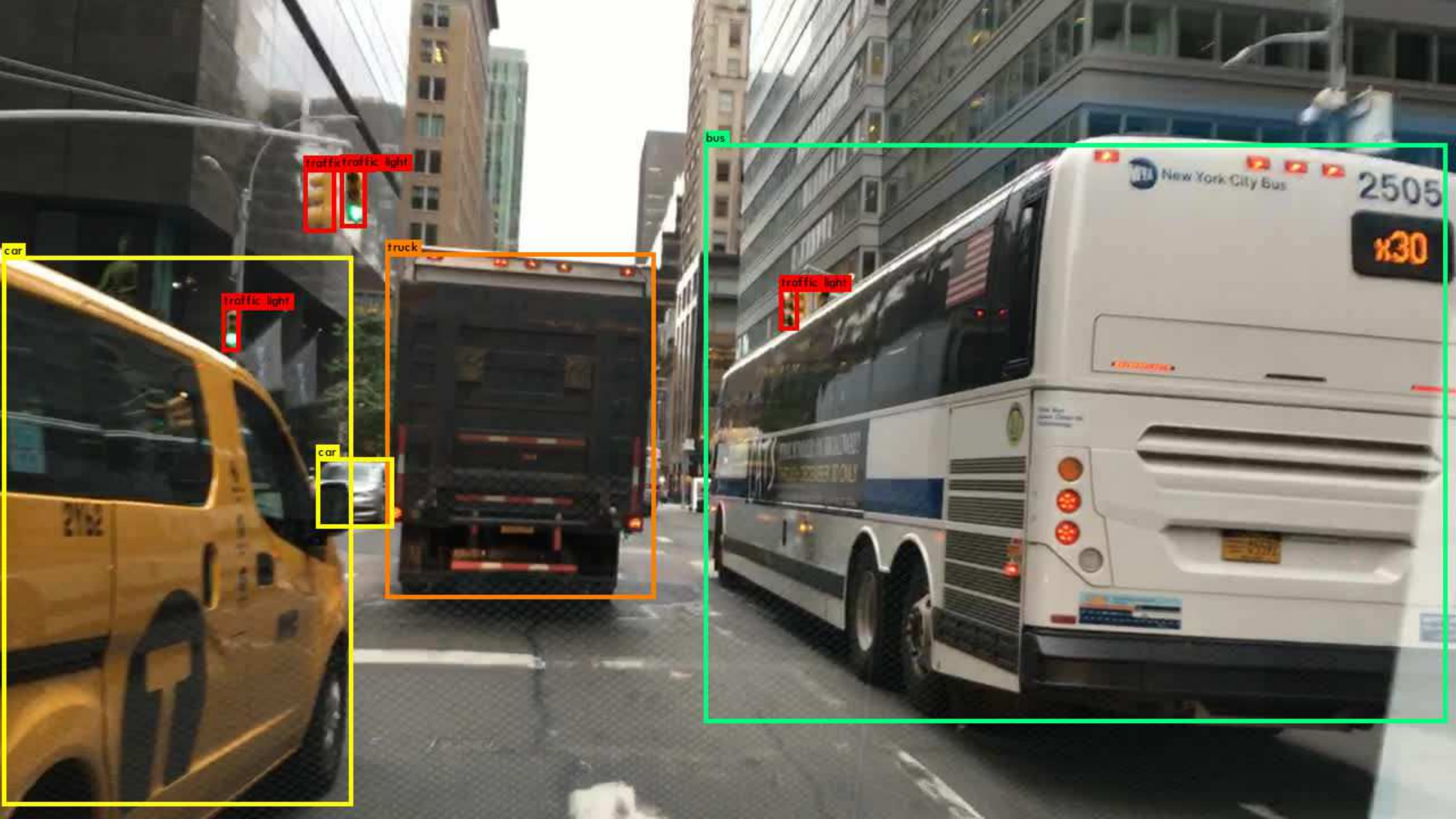}	
		\caption{Detection results of the baseline and proposed algorithms on the BDD test set. The first and second rows show the detection results of YOLOv3 and Gaussian YOLOv3, respectively, and each color is related to a particular object class.}
		\label{fig:bdd_det_result}
	\end{figure*}

	For a visual evaluation of Gaussian YOLOv3, Figures~\ref{fig:kitti_det_result} and~\ref{fig:bdd_det_result} show the detection examples of the baseline and Gaussian YOLOv3 for the KITTI validation set and the BDD test set, respectively. The detection TH is 0.5, which is the default test TH of YOLOv3. The results in the first row of Figure~\ref{fig:kitti_det_result} and in the first column of Figure~\ref{fig:bdd_det_result} show that Gaussian YOLOv3 can detect objects that YOLOv3 cannot find, thereby increasing its TP. These positive results are obtained because the Gaussian modeling and loss function reconstruction of YOLOv3 proposed in this paper can provide a loss attenuation effect in the learning process, so that the learning accuracy for bbox can be improved, which enhances the performance of objectness. Next, the results in the second row of Figure~\ref{fig:kitti_det_result} and in the second column of Figure~\ref{fig:bdd_det_result} show that Gaussian YOLOv3 can complement incorrect object detection results found by YOLOv3. In addition, the results in the third row of Figure~\ref{fig:kitti_det_result} and in the third column of Figure~\ref{fig:bdd_det_result} show that Gaussian YOLOv3 can accurately detect bbox of object inaccurately detected by YOLOv3. Based on these results, Gaussian YOLOv3 can significantly reduce the FP and increase the TP, and consequently, the driving stability and efficiency are improved and fatal accidents can be prevented.
	\begin{table}[t!]
		\centering
		\resizebox{0.41\textwidth}{!}
		{
			\begin{tabular}{cccc}
				\hline\hline
				& YOLOv3  & \begin{tabular}[c]{@{}c@{}}Gaussian\\ YOLOv3\end{tabular} & \begin{tabular}[c]{@{}c@{}}Variation\\ rate (\%)\end{tabular} \\ \hline
				KITTI validation set &         &                                                           &                                                               \\
				\# of FP             & 1,681   & 985                                                       & -41.40                                                        \\
				\# of TP             & 13,575  & 14,560                                                    & +7.26                                                         \\
				\# of GT             & 17,607  & 17,607                                                    & 0                                                             \\ \hline
				BDD validation set   &         &                                                           &                                                               \\
				\# of FP             & 86,380  & 51,296                                                    & -40.62                                                        \\
				\# of TP             & 57,261  & 59,724                                                    & +4.30                                                         \\
				\# of GT             & 185,578 & 185,578                                                   & 0                                                             \\ \hline\hline
			\end{tabular}
		}
		\caption{Numerical evaluation of FP and TP.}
		\label{table_fp_tp_num}
	\end{table}
	
	For a numerical evaluation of the FP and TP of Gaussian YOLOv3, Table~\ref{table_fp_tp_num} shows the numbers of FPs and TPs for the baseline and Gaussian YOLOv3. The detection TH is the same as the mentioned before. The KITTI and BDD validation sets are used to calculate the FP and TP because the GT is provided in the validation set. For more accurate measurements, the FP and TP of the two datasets are calculated using the official evaluation code of BDD because the KITTI official evaluation method does not count the FP when bbox is within a certain size. For the KITTI and BDD validation sets, Gaussian YOLOv3 reduces the FP by 41.40\% and 40.62\%, respectively, compared to YOLOv3. In addition, it increases the TP by 7.26\% and 4.3\%, respectively. It should be noted that the reduction in the FP prevents unnecessary unexpected braking, and the increase in the TP prevents fatal accidents from object detection errors. In conclusion, Gaussian YOLOv3 shows a better performance than YOLOv3 for both the FP and TP related to the safety of autonomous vehicles. Based on the results described in Sections 4.1, 4.2, and 4.3, the proposed algorithm outperforms previous studies and is most suitable for autonomous driving applications.
	\section{Conclusion}
	A high accuracy and real-time detection speed of an object detection algorithm are extremely important for the safety and real-time control of autonomous vehicles. Various studies related to camera-based autonomous driving have been conducted, but are unsatisfactory based on a trade-off between the accuracy and operation speed. For this reason, this paper proposes an object detection algorithm that achieves the best trade-off between accuracy and speed for autonomous driving. Through Gaussian modeling, loss function reconstruction, and the utilization of localization uncertainty, the proposed algorithm improves the accuracy, increases the TP, and significantly reduces the FP, while maintaining the real-time capability. Compared to the baseline, the proposed Gaussian YOLOv3 algorithm improves the mAP by 3.09 and 3.5 for the KITTI and BDD datasets, respectively. Furthermore, because the proposed algorithm has a higher accuracy than the previous studies with a similar fps, the proposed algorithm is excellent in terms of the trade-off between accuracy and detection speed. As a result, the proposed algorithm can significantly improve the camera-based object detection system for autonomous driving, and is consequently expected to contribute significantly to the wide use of autonomous driving applications.
	
	\section*{Acknowledgement}
	This work was supported by the National Research Foundation of Korea (NRF) grant funded by the Korea government (MSIT) (No. 2019R1F1A1057530) and "The Project of Industrial Technology Innovation" through the Ministry of Trade, Industry and Energy (MOTIE) (10082585,2017).
	
	\clearpage
	\newpage
	
	{\small
		\bibliographystyle{ieee_fullname}
		\bibliography{egbib}
	}
	
\end{document}